\documentclass[sigconf,screen]{acmart}

\usepackage{algorithm}
\usepackage{algorithmicx}
\usepackage{algpseudocode}
\usepackage{array}
\usepackage{balance}
\usepackage{booktabs}
\usepackage{colortbl}
\usepackage{enumitem}
\usepackage{fontawesome}
\usepackage{geometry}
\usepackage{graphicx}
\usepackage{makecell}
\usepackage{mdframed}
\usepackage{multirow}
\usepackage{subfigure}
\usepackage{tcolorbox}
\usepackage{threeparttable}
\usepackage{ulem}
\usepackage{url}
\usepackage{wrapfig}
\usepackage{xcolor}

\AtBeginDocument{%
  \providecommand\BibTeX{{%
    \normalfont B\kern-0.5em{\scshape i\kern-0.25em b}\kern-0.8em\TeX}}}

\setcopyright{acmlicensed}
\acmDOI{10.1145/3650212.3652122}
\acmYear{2024}
\copyrightyear{2024}
\acmSubmissionID{issta24main-p445-p}
\acmISBN{979-8-4007-0612-7/24/09}
\acmConference[ISSTA '24]{Proceedings of the 33rd ACM SIGSOFT International Symposium on Software Testing and Analysis}{September 16--20, 2024}{Vienna, Austria}
\acmBooktitle{Proceedings of the 33rd ACM SIGSOFT International Symposium on Software Testing and Analysis (ISSTA '24), September 16--20, 2024, Vienna, Austria}
\begin{document}

\title{A Large-Scale Empirical Study on Improving the Fairness of Image Classification Models}


\author{Junjie Yang}
\affiliation{
\department{College of Intelligence and Computing}
    \institution{Tianjin University}
    \country{Tianjin, China}
}
\email{jjyang@tju.edu.cn}

\author{Jiajun Jiang}
\authornote{Corresponding author.}
\affiliation{
\department{College of Intelligence and Computing}
    \institution{Tianjin University}
    \country{Tianjin, China}
}
\email{jiangjiajun@tju.edu.cn}

\author{Zeyu Sun}
\affiliation{
    \institution{Science \& Technology on Integrated Information System Laboratory, Institute of Software, Chinese Academy of Sciences}
    \country{Beijing, China}
}
\email{zeyu.zys@gmail.com}

\author{Junjie Chen}
\affiliation{
\department{College of Intelligence and Computing}
    \institution{Tianjin University}
    \country{Tianjin, China}
}
\email{junjiechen@tju.edu.cn}


\newcommand{\link}[1]{\textbf{https://github.com/junjie1003/DL-Fairness-Study}}

\newcommand{\junjie}[1]{{\color{pink}[Junjie: #1]}}
\newcommand{\jiajun}[1]{{\color{red}[Jiajun: #1]}}
\newcommand{\zeyu}[1]{{\color{blue}[Zeyu: #1]}}

\newcommand{\hl}[1]{\cellcolor{gray!25}\textbf{#1}}
\newcommand{\ul}[1]{\color{blue}\uline{\color{black}\textbf{#1}}}


\newcounter{finding}
\newcommand{\finding}[1]{\refstepcounter{finding}
    \vspace{0.5mm}
    \begin{mdframed}[linecolor=gray,roundcorner=12pt,backgroundcolor=gray!15,linewidth=3pt,innerleftmargin=2pt, leftmargin=0cm,rightmargin=0cm,topline=false,bottomline=false,rightline = false]
        \textbf{Finding \arabic{finding}:} #1
    \end{mdframed}
    \vspace{0.5mm}
}

\begin{abstract}
Fairness has been a critical issue that affects the adoption of deep learning models in real practice. To improve model fairness, many existing methods have been proposed and evaluated to be effective in their own contexts. However, there is still no systematic evaluation among them for a comprehensive comparison under the same context, which makes it hard to understand the performance distinction among them, hindering the research progress and practical adoption of them. To fill this gap, this paper endeavours to conduct the first large-scale empirical study to comprehensively compare the performance of existing state-of-the-art fairness improving techniques. Specifically, we target the widely-used application scenario of image classification, and utilized three different datasets and five commonly-used performance metrics to assess in total 13 methods from diverse categories. Our findings reveal substantial variations in the performance of each method across different datasets and sensitive attributes, indicating over-fitting on specific datasets by many existing methods. Furthermore, different fairness evaluation metrics, due to their distinct focuses, yield significantly different assessment results. Overall, we observe that pre-processing methods and in-processing methods outperform post-processing methods, with pre-processing methods exhibiting the best performance. Our empirical study offers comprehensive recommendations for enhancing fairness in deep learning models. We approach the problem from multiple dimensions, aiming to provide a uniform evaluation platform and inspire researchers to explore more effective fairness solutions via a set of implications.
\end{abstract}






\maketitle

\section{Introduction}
\label{sec:intro}

In recent years, Artificial Intelligence (AI) systems and Deep Learning (DL) models have attracted significant attention for their remarkable performance across diverse application domains, such as image processing~\cite{DBLP:journals/sigpro/Saarinen94,DBLP:books/lib/GonzalezW08}, machine translation~\cite{DBLP:journals/coling/BrownPPM94,DBLP:conf/mtsummit/Koehn05,DBLP:conf/acl/Chiang05,DBLP:conf/acl/KoehnHBCFBCSMZDBCH07} and software engineering~\cite{DBLP:journals/computer/Brooks87,DBLP:conf/icse/Royce87,yang2023code,DBLP:conf/icse/YouWCLL23}. 
However, the increasing adoption of DL models in real practice also poses new challenges to them. In particular, they have displayed troubling discriminatory tendencies in certain applications, such as elevated error rates, when confronted with specific groups or populations defined by attributes deemed \textit{protected} or \textit{sensitive}~\cite{DBLP:conf/fat/FriedlerSVCHR19,DBLP:journals/csur/MehrabiMSLG21,DBLP:conf/sigsoft/Zhang022,DBLP:conf/sigsoft/BiswasR20,DBLP:conf/sigsoft/ChenZSH22,DBLP:journals/corr/abs-1905-05786}, including factors like race, gender, and age, etc., causing ethical or even safety issues that will largely affect the reliability and usability of them in real practice. For example, 
AI judge~\cite{steinitz2014case,green2019principles,huq2018racial,xu2022human} may exhibit biased decision-making, leading to unequal treatment of individuals based on their sensitive attributes, such as race. This bias can result in disparities in sentencing or legal outcomes, raising concerns about fairness and justice within the legal system. Similarly, AI-powered hiring~\cite{palshikar2018automatic,harirchian2022ai} and loan systems~\cite{galindo2000credit,wang2011comparative} tend to favor applicants of a particular gender or race. 
In other words, the manifestation of unfair behaviors in AI systems and DL models has raised profound ethical concerns, transforming fairness into an essential ethical and often legally mandated prerequisite for the widespread adoption and utilization of DL models across various real-world contexts.

To improve the fairness of DL models, many approaches have been proposed in the last decades, particularly in the application domain of image classification, which has been widely adopted in diverse real-world applications. 
Typically, existing approaches can be classified into three categories according to the stage where de-biasing is applied, i.e., pre-processing, in-processing, and post-processing. Pre-processing techniques~\cite{DBLP:journals/nn/BudaMM18,DBLP:journals/ida/JapkowiczS02,DBLP:journals/corr/abs-2209-15605,DBLP:journals/corr/abs-2211-00168,DBLP:journals/corr/abs-2005-04345,DBLP:conf/cvpr/RamaswamyKR21,DBLP:conf/cvpr/WangQKGNHR20} endeavor to mitigate model biases by optimizing the training data, such as performing data transformation ~\cite{DBLP:journals/corr/abs-2211-00168}, augmentation ~\cite{DBLP:conf/cvpr/RamaswamyKR21}, sampling ~\cite{DBLP:journals/nn/BudaMM18,DBLP:journals/ida/JapkowiczS02,DBLP:journals/corr/abs-2005-04345,DBLP:conf/cvpr/WangQKGNHR20,DBLP:journals/corr/abs-2209-15605} and so on. In contrast, in-processing techniques~\cite{DBLP:conf/iccv/TzengHDS15,DBLP:conf/nips/HongY21,DBLP:journals/corr/abs-2304-14252,DBLP:conf/cvpr/JungLPM21,DBLP:conf/eccv/KehrenbergBTQ20} dedicate to improve the model fairness by optimizing the model training process, such as updating the training loss~\cite{DBLP:conf/nips/HongY21,DBLP:journals/corr/abs-2304-14252} or model structures~\cite{DBLP:conf/cvpr/JungLPM21,DBLP:conf/eccv/KehrenbergBTQ20}.
Different from the former two categories, post-processing techniques~\cite{DBLP:conf/aies/KimGZ19} aim to make the deployed models produce fairer outputs without changing the model structures and parameters, such as optimizing the given inputs~\cite{DBLP:conf/nips/ZhangZ0F00C22,DBLP:conf/cvpr/WangDXZCW022}.
Since they do not change the original models, they can be flexibly applied to improve the fairness of deployed models in real practice.

Although these existing approaches have been evaluated to be effective in their own contexts, as will be presented in Section~\ref{sec:limitation}, they were often evaluated over different datasets and utilized inconsistent metrics for measuring the performance of the proposed methods, making it hard to obtain a uniform conclusion about the performance distinction among them over different application domains. In other words, there still lacks a comprehensive study to systematically compare the performance of different approaches under the same setup and thus provide in-depth analysis and accurate guidance for the better utilization of them in practice.

To fill this gap, we conducted the first extensive empirical study to comprehensively compare the performance of existing methods for improving the fairness of deep learning models. Specifically, we studied 13 state-of-the-art existing techniques for improving the fairness of deep learning models as the representatives, all of which are from the most recent research. To provide a comprehensive and fair comparison, we adopted three widely-used benchmarks that involve diverse image classification tasks related to different sensitive attributes. Moreover, we employed all five widely-used fairness metrics and two accuracy metrics in the result analysis for a consistent measurement, which is the most comprehensive study on model fairness as far as we are aware. According to the empirical results, we have summarized a set of findings and implications that can be valuable to guide future research and the utilization of them in real practice. We highlight partial key findings as follows:

\begin{enumerate}[leftmargin=*]
    \item While the best-performing fairness improvement method does not exist, pre-processing and in-processing methods significantly outperform post-processing methods.
    \item Existing methods tend to be \textit{insensitive} to different evaluation metrics, and thus a subset of the metrics can be representative in future studies.
    \item Existing methods are \textit{sensitive} to different experimental datasets, and thus the conclusions drawn from a certain dataset may not be generalizable to others.
\end{enumerate}

To sum up, our work makes the following major contributions:

\begin{itemize}[leftmargin=*]
    \item We conducted \textbf{the first large-scale empirical study} to comprehensively evaluate the performance of 13 state-of-the-art techniques that aim to improve DL models' fairness.
    \item We summarized \textbf{a set of findings and implications} by systematically analyzing and comparing the performance of different techniques under a uniform experimental setup. 
    \item We re-implemented some of the state-of-the-art techniques and built \textbf{a uniform evaluation platform} of DL fairness techniques, which can facilitate the replication and comparison for future research in this research area. We have published all our experimental data.
\end{itemize}


\section{Background}
\label{sec:back}


Firstly, we would like to clarify the problem and concept of \textit{fairness issues} in deep learning models. Furthermore, we will also conduct a succinct literature review of the most recent research on fairness improvement and identify the limitations of them, which motivates the necessity and significance of this study.

\subsection{Fairness Issues}

Over the years, both researchers and practitioners have introduced and investigated various fairness definitions ~\cite{DBLP:journals/corr/abs-2106-00467,DBLP:journals/csur/MehrabiMSLG21,DBLP:journals/corr/abs-2207-10223}. These definitions typically fall into two broad categories: \textit{individual fairness} and \textit{group fairness}. Individual fairness requires that software produces similar predictive outcomes for individuals with similar characteristics, while group fairness mandates equitable treatment of different demographic groups. In particular, fairness is always closely related to the concept of \textit{sensitive} or \textit{protected} attributes, which 
represent characteristics demanding protection against discriminatory practices, such as gender and race.
In particular, if the assigned value of a sensitive (or protected) attribute is considered advantageous or beneficial in a particular context, we call it a \textit{privileged value}. For example, in a job application process~\cite{van2019marketing,van2021job,rodney2019artificial}, \textit{male} would be a privileged value for the attribute of gender since man may tend to be preferred over women for certain roles.
Therefore, given the sensitive attribute (e.g., gender), the input instances of the deep learning models can be typically divided into two distinct groups: a privileged group that associates instances with privileged values (e.g., male) and an unprivileged group that associates instances with unprivileged values (e.g., female). 
A \textit{fair} DL model should produce similar or even equivalent prediction results over different groups. 

In fact, software fairness issues (i.e., unfair software predictions)~\cite{DBLP:conf/re/FinkelsteinHMRZ08} have been a growing area of concern in both the Software Engineering (SE) and Deep Learning (DL) research communities.
For example, they are typically referred to as \textit{fairness defects} ~\cite{DBLP:conf/sigsoft/BrunM18} or \textit{fairness bugs} ~\cite{DBLP:conf/sigsoft/ChenZSH22} in the SE community. In this study, we refer to the unfair predictions produced by DL models as \textit{fairness issues} by following existing work~\cite{DBLP:conf/re/FinkelsteinHMRZ08}, indicating a discordance between existing and expected fairness criteria. 
Many testing~\cite{DBLP:journals/corr/abs-2207-10223,DBLP:conf/sigsoft/GalhotraBM17,DBLP:conf/icst/PatelCLKK22,DBLP:conf/issta/GuoLWL0HZ023,DBLP:conf/sigsoft/AggarwalLNDS19,sun2020automatic,DBLP:journals/tse/SoremekunUC22,sun2022improving,DBLP:conf/sigsoft/WangYCLZ20,DBLP:conf/kbse/Shen0ZW0T22} and repair studies~\cite{DBLP:conf/sigsoft/NguyenBR23,DBLP:conf/sigmod/SalimiRHS19,DBLP:conf/icse/Sun0PS22} aim to automatically identify and fix such issues. 
In this study, we direct our attention specifically to the techniques that aim to resolve the fairness issues in image classification models since they have been already widely deployed in practice as explained in the introduction. In addition, this study focuses on the \textit{group fairness} of DL models.

\begin{table*}[t]
    \caption{Summary of fairness improving methods.}
    \label{tab:works}
    \centering
    \resizebox{\textwidth}{!}{
        \begin{tabular}{c|c|lll}
            \toprule
            \textbf{Category}                                                           & \textbf{Data Type} & \textbf{Methods}                                                                                                                                                                                                                                                                                                                                                                                                                                                            & \textbf{Sensitive Attributes}                                                                                           & \textbf{Fairness Metrics}                                              \\
            \midrule
            \multicolumn{1}{c|}{\multirow{4}{*}{\makecell[c]{Pre-processing}}} & Tabular   & \makecell[l]{OP ~\cite{DBLP:conf/nips/CalmonWVRV17}, Fair-SMOTE ~\cite{DBLP:conf/sigsoft/ChakrabortyMM21}, RW ~\cite{DBLP:journals/kais/KamiranC11}, DIR ~\cite{DBLP:conf/kdd/FeldmanFMSV15},\\ LTDD ~\cite{DBLP:conf/icse/LiMC0WZX22}, FairPreprocessing ~\cite{DBLP:conf/sigsoft/BiswasR21}, Fairway ~\cite{DBLP:conf/sigsoft/ChakrabortyM0M20}                                                                                                              } & \makecell[l]{gender, age, race, black}                                                                         & \makecell[l]{AAOD, EOD, BER,\\ SPD, DI, FPR, ERD}           \\
            \cline{2-5}
            \multicolumn{1}{c|}{}                                              & Image     & \makecell[l]{FairHSIC ~\cite{DBLP:conf/cvpr/QuadriantoST19}, \textbf{US} ~\cite{DBLP:journals/corr/abs-2005-04345}, \textbf{OS} ~\cite{DBLP:conf/cvpr/WangQKGNHR20}, \textbf{UW} ~\cite{DBLP:journals/corr/abs-2209-15605},\\ GAN-debiasing ~\cite{DBLP:conf/cvpr/RamaswamyKR21}, CGL~\cite{DBLP:conf/cvpr/JungCM22}, \textbf{BM} ~\cite{DBLP:journals/corr/abs-2209-15605}                                                                                 } & \makecell[l]{gender, background, \\ age, race, color/grayscale}                                                & \makecell[l]{EOD, DEO, \\ BA, BC, KL}                         \\
            \hline
            \multirow{5}{*}{\makecell[c]{In-processing}}                       & Tabular   & \makecell[l]{META ~\cite{DBLP:conf/fat/CelisHKV19}, AD ~\cite{DBLP:conf/aies/ZhangLM18}, GR ~\cite{DBLP:conf/icml/AgarwalBD0W18}, MAAT ~\cite{DBLP:conf/sigsoft/ChenZSH22},\\ PR ~\cite{DBLP:conf/pkdd/KamishimaAAS12}, FMT ~\cite{Jiang2023APF}, CARE ~\cite{DBLP:conf/icse/Sun0PS22}, Parfait-ML ~\cite{DBLP:conf/icse/Tizpaz-NiariKT022}                                                                                                                    } & \makecell[l]{gender, age, race}                                                                                & \makecell[l]{NMI, PI/MI, NPI, UEI,\\ CVS, SPD, AAOD, EOD} \\
            \cline{2-5}
                                                                               & Image     & \makecell[l]{\textbf{Adv} ~\cite{DBLP:conf/cvpr/WangQKGNHR20}, \textbf{DI} ~\cite{DBLP:conf/cvpr/WangQKGNHR20}, \textbf{BC+BB} ~\cite{DBLP:conf/nips/HongY21}, \\ \textbf{MFD} ~\cite{DBLP:conf/cvpr/JungLPM21}, FairBatch ~\cite{DBLP:conf/iclr/Roh0WS21}, FSCL ~\cite{DBLP:conf/cvpr/ParkLLHKB22}, \\ FairGRAPE ~\cite{DBLP:conf/eccv/LinKJ22}, \textbf{FDR} ~\cite{DBLP:journals/corr/abs-2304-03935}, \textbf{FLAC} ~\cite{DBLP:journals/corr/abs-2304-14252}} & \makecell[l]{background color, \\ color (bright/dark), \\ age, race, texture bias, \\ gender, color/grayscale} & \makecell[l]{EOD, BC, WA, \\ DEO, SPD, AED}                   \\
            \hline
            \multirow{3}{*}{\makecell[c]{Post-processing}}                     & Tabular   & \makecell[l]{EO ~\cite{DBLP:conf/nips/HardtPNS16}, CEO ~\cite{DBLP:conf/nips/PleissRWKW17}, ROC ~\cite{DBLP:conf/icdm/KamiranKZ12}, Fax-AI ~\cite{DBLP:conf/fat/GrabowiczPM22}                                                                                                                                                                                                                                                                                 } & \makecell[l]{gender, age, race}                                                                                & \makecell[l]{DEO, EOD}                                        \\
            \cline{2-5}
                                                                               & Image     & \makecell[l]{\textbf{FairReprogram} ~\cite{DBLP:conf/nips/ZhangZ0F00C22}, \\ Multiaccuracy ~\cite{DBLP:conf/aies/KimGZ19}, \textbf{FAAP} ~\cite{DBLP:conf/cvpr/WangDXZCW022}}                                                                                                                                                                                                                                                                                      & \makecell[l]{gender}                                                                                           & \makecell[l]{DEO, SPD}                                        \\
            \bottomrule
        \end{tabular}
    }
\end{table*}

\subsection{Limitations in Existing Studies}
\label{sec:limitation}

In recent years, there has been a growing focus among researchers on fairness issues, leading to various of fairness improvement methods being proposed. In order to gain a deeper understanding of existing studies, we have conducted a succinct literature review of relevant papers published recently in prominent conferences and journals in the Software Engineering (SE) and Artificial Intelligence (AI) domains, including ICSE, ESEC/FSE, TOSEM, AIES, ICLR, ECCV, CVPR, NeurIPS, etc.
Table~\ref{tab:works} presents the details of these methods that aim to improve DL models' fairness.

In this table, besides the essential information of the method (e.g., method names and categories), we further classify the methods into \textit{Tabular} and \textit{Image} based on the type of dataset they are designed for. Additionally, we display the union sets of sensitive attributes and fairness metrics adopted by all methods within each category. Please note that there are also different classification methods~\cite{caton2020fairness}. In this paper we adopt the most widely-used one ~\cite{DBLP:journals/csur/MehrabiMSLG21,DBLP:conf/sigsoft/Zhang022,DBLP:conf/sigsoft/BiswasR20} for easy understanding. After a careful examination, we have the following two major observations regarding the limitations in existing studies.

\begin{description}[leftmargin=*]
    \item[Limitation 1] \textbf{(Incomplete or Unrepresentative Datasets):} Most of existing methods were evaluated on only one or two datasets, with a strong preference towards facial datasets. While face datasets offer advantages such as ease of acquisition and mature application scenarios, the real applications of such techniques are not limited to the classification based on the facial images, such as scenes or objects. However, only a few methods have been evaluated on a sufficiently diverse set of datasets and task scenarios, which is indeed inadequate. Furthermore, many methods use datasets that are rarely employed by other works, e.g., DiF~\cite{DBLP:journals/corr/abs-1901-10436} and Waterbirds~\cite{DBLP:journals/corr/abs-1911-08731}, lacking the representation and making it harder to horizontally compare the performance of different techniques. In particular, studies in the SE community primarily focus on tabular datasets, which are too simple to assess the generalizability of the proposed method for a broader range of applications. In contrast, image data is more complex, and the models used are also more sophisticated. Fairness research on image data is more widespread and holds high research value. However, there still lacks a comprehensive evaluation of existing approaches over image datasets aiming to provide implications for future research.


    \item[Limitation 2] \textbf{(Inconsistent Fairness Evaluation Metrics):} Most of the existing studies only adopted one or two metrics, which may not offer a complete assessment of model fairness. Moreover, some works introduce unique fairness metrics~\cite{DBLP:conf/aies/KimGZ19,DBLP:conf/eccv/LinKJ22} tailored to their specific task scenarios, but these metrics are rarely cited or adopted by other research, raising questions about the reliability and generalizability of these metrics. In particular, the different metrics adopted by existing studies further hinder the comparison of different techniques.
\end{description}


The limitations above pose an emergent need to conduct a large-scale empirical study for comprehensively comparing their performance under a uniform context, which motivates this work. In particular, to address these limitations, we have replicated the experimental results of many existing approaches over their original datasets and then further evaluated their effectiveness over three different datasets involving both facial and non-facial images (i.e., objects). Furthermore, we have employed all the widely-used performance metrics (i.e., five fairness metrics and two accuracy metrics) for systematically demonstrating the effectiveness of them.
In this way, we can better understand the performance of existing techniques and derive valuable insights for bias mitigation in deep learning models. Ultimately, our aspiration is to provide meaningful guidance that will facilitate more effective explorations of fairness issues in the realm of image fairness.




\section{Studied Methods in This Study}
\label{sec:methods}

As summarized in Table~\ref{tab:works}, many novel methods have been published in a wide range of conferences and journals, aiming to enhance the fairness of neural networks.
The latest research~\cite{chen2024fairness} has explored the performance of existing methods that focus on tabular datasets, while those methods targeting image datasets have not been systematically investigated yet. Therefore, our study will primarily focus on fairness improving methods for image datasets. Specifically, we follow the selection criteria below to select the methods ultimately studied in our study.

\begin{description}[leftmargin=*]
    \item[Criterion 1] \textbf{(Open-source or Reproducible):} To ensure the correctness and reliability of our study, we have tried to reproduce the  results of the selected methods before conducting our experiment. Then, we selected the methods that are open-source or can be replicated via our re-implementation according to the reported configurations, while ignored those that cannot replicate. In particular, we appreciate the corresponding authors that assisted us in re-implementing and reproducing their results.

    \item[Criterion 2] \textbf{(Covering Different Categories):} To ensure the comprehensiveness of our study, we have selected methods for investigation in all three categories: pre-processing, in-processing, and post-processing, ensuring coverage across all categories.

    \item[Criterion 3] \textbf{(Latest Studies):} To ensure the effectiveness and representativeness of our study, we prefer to select the methods that were most recently proposed since they usually can achieve much better performance than the previous ones.

\end{description}

As a result, 13 state-of-the-art methods (including two variants of FR~\cite{DBLP:conf/nips/ZhangZ0F00C22}) were included, involving all the three categories, i.e., pre-processing, in-processing, and post-processing. We also emphasized these methods in \textbf{bold} in Table~\ref{tab:works} and introduced them as follows.

\subsection{Pre-processing}
\label{sec:pre-proc}

Pre-processing methods, as a general practice, are primarily concerned with reducing discrimination and bias within the training data, with the ultimate objective of improving the fairness of the resulting trained model. These techniques involve data transformations and augmentations designed to mitigate inherent forms of discrimination. Pre-processing strategies are employed when the algorithm has the capacity to modify the training data to mitigate bias. From the plethora of available pre-processing methods ~\cite{DBLP:journals/nn/BudaMM18,DBLP:journals/ida/JapkowiczS02,DBLP:conf/cvpr/WangQKGNHR20,DBLP:conf/icml/ByrdL19,DBLP:journals/corr/abs-2209-15605,DBLP:conf/cvpr/JungCM22,DBLP:journals/corr/abs-2211-00168,DBLP:conf/cvpr/QuadriantoST19,DBLP:journals/corr/abs-2005-04345}, we specifically concentrate on the following four approaches as the representative based on the selection criteria.

\begin{itemize}[leftmargin=*]
    \item \textbf{Undersampling (US)} ~\cite{DBLP:journals/nn/BudaMM18,DBLP:journals/ida/JapkowiczS02,DBLP:journals/corr/abs-2005-04345} aims at achieving class balance by reducing samples from the majority class. It has also been extended to tackle bias by selectively dropping samples from specific subgroups, which include samples that share common class and bias attributes. This approach effectively balances the sizes of these subgroups. However, a notable challenge arises as the model's exposure to samples becomes constrained by the size of the smallest subgroup.
    \item \textbf{Oversampling (OS)} ~\cite{DBLP:conf/cvpr/WangQKGNHR20} is employed to rectify class imbalance by duplicating samples, thereby equalizing class sizes. This approach has been adapted to effectively address bias by ensuring balance within sensitive subgroups. In practice, we ascertain the size of the largest subgroup and then replicate samples in other subgroups proportionally. However, it's important to exercise caution, as excessive sample duplication can result in overfitting, especially when dealing with highly parameterized models like deep neural networks.
    \item \textbf{Upweighting (UW)} ~\cite{DBLP:conf/icml/ByrdL19,DBLP:journals/corr/abs-2209-15605} balances the impact of different samples on the loss function by scaling the loss value with the inverse of the sample's class frequency. Therefore, this technique can mitigate bias across different subgroups rather than individual classes. We classify it into the pre-processing category since it depends on the statistics of the training data. Nevertheless, Upweighting possesses a notable limitation—it can exhibit instability when used in conjunction with stochastic gradient descent ~\cite{DBLP:conf/iclr/AnYZ21}.
    \item \textbf{Bias Mimicking (BM)} ~\cite{DBLP:journals/corr/abs-2209-15605} was proposed to address dataset bias in Visual Recognition tasks. It achieves statistical independence between bias and class labels by emulating bias distributions across classes. This method involves subsampling the dataset into diverse distributions per class, striking a balance between simplicity and performance enhancement, effectively bridging the gap between sampling and non-sampling methods.
\end{itemize}

\subsection{In-processing}
\label{sec:in-proc}

In-processing methods endeavor to modify state-of-the-art learning algorithms during the model training process to eliminate discrimination. This is primarily achieved through the integration of adjustments into the objective function or the imposition of constraints on the learning process, provided that the model's learning procedure is adaptable to such modifications. These techniques employ various means to adjust the model and mitigate bias in its predictions. In our study, we employ the following six methods.

\begin{itemize}[leftmargin=*]
    \item \textbf{Adversarial Training (Adv)} with uniform confusion ~\cite{DBLP:conf/iccv/TzengHDS15,DBLP:journals/corr/abs-2209-15605} introduces an adversarial loss to induce a randomized feature representation of target labels and bias groups within the model. It introduces an innovative CNN architecture specifically crafted to leverage unlabeled and sparsely labeled data in the target domain. Additionally, the architecture incorporates domain confusion and softmax cross-entropy losses for model training.

    \item \textbf{Domain Independent Training (DI)} ~\cite{DBLP:conf/cvpr/WangQKGNHR20,DBLP:journals/corr/abs-2209-15605} involves the utilization of extra prediction heads for each bias group. In response to the issue of discriminative models learning unnecessary domain-specific class boundaries, DI promotes the adoption of distinct domain-specific classifiers. Nonetheless, training separate classifiers diminishes data exposure. To address this concern, DI proposes a shared feature representation alongside an ensemble of classifiers. It recommends direct reasoning on class boundaries of domains, effectively eradicating class-domain correlations.

    \item \textbf{Bias-Contrastive and Bias-Balanced Learning (BC+BB)} ~\cite{DBLP:conf/nips/HongY21,DBLP:journals/corr/abs-2209-15605} introduces an innovative method to mitigate bias by combining two distinct losses. The Bias-Contrastive (BC) loss incorporates contrastive learning, leveraging bias labels to effectively mitigate bias. Furthermore, the Bias-Balanced (BB) regression loss enhances debiasing performance by optimizing the model to achieve a uniform target-bias correlation distribution.

    \item \textbf{FLAC} ~\cite{DBLP:journals/corr/abs-2304-14252} mitigates bias in DL models by reducing mutual information between model-extracted features and a protected attribute. It utilizes a sampling strategy to highlight underrepresented data and transforms fair representation learning into a probability matching task using representations obtained from a bias-capturing classifier. This approach disentangles the target representation from bias and protected attributes, ultimately leading to fairer outcomes.

    \item \textbf{MMD-based Fair Distillation (MFD)} ~\cite{DBLP:conf/cvpr/JungLPM21} combines feature distillation with maximum mean discrepancy (MMD) to enhance both prediction accuracy and fairness simultaneously. The theoretical foundation demonstrates that its MMD-based regularization fosters fairness by aligning group-conditioned features of the student model across sensitive groups and aligning them with the teacher model's group-averaged features, resulting in improved accuracy as well.

    \item \textbf{Fair Deep Feature Reweighting (FDR)} ~\cite{DBLP:journals/corr/abs-2304-03935} is a simple and straightforward method that improves the model fairness by fine-tuning, where the fairness metrics are incorporated when computing the loss for model updating. In particular, only the last-layer will be updated during the tuning process.
\end{itemize}

\subsection{Post-processing}
\label{sec:post-proc}

Post-processing methods are employed to improve fairness in deep learning software by modifying prediction outcomes. These methods are applied after the model training process. When dealing with models as black boxes without the capability to modify training data or algorithms, post-processing becomes essential and easy to use in practice. In this study, we adopted three methods in this category, including two variants of the FairReprogram.

\begin{itemize}[leftmargin=*]
    \item \textbf{FairReprogram (FR)} ~\cite{DBLP:conf/nips/ZhangZ0F00C22} leverages model reprogramming techniques to address scenarios where it's not feasible to modify existing models. Instead, it appends a set of perturbations known as the "fairness trigger" to the input data, optimizing it in a min-max framework to align with fairness criteria. In particular, FR applies two methodologies to append the trigger, corresponding to its 2 variants, named \textbf{FR-P} (appending the trigger like a patch to the original image) and \textbf{FR-B} (appending the trigger at the border of the original image). We applied both of them in our study.
    Specifically, they perturb the inputs by adding a constant global vector/feature, which is tailored to improve fairness while keeping the deep learning model unchanged. FairReprogram is a versatile framework applicable to various tasks and domains, making it a generic solution.
    \item \textbf{Fairness-Aware Adversarial Perturbation (FAAP)} ~\cite{DBLP:conf/cvpr/WangDXZCW022} mitigates unfairness in deployed deep models by updating the given input without modifying model parameters or structures. Specifically, it achieves this by learning to perturb input data, effectively blinding deployed models on fairness-related attributes such as gender or ethnicity. FAAP employs a discriminator-generator adversarial framework to ensure that protected attributes are not correlated with model predictions, making it flexible and practical to addressing fairness bugs in real-world AI systems.
\end{itemize}

\section{Experimental Setup}
\label{sec:setup}


\subsection{Dataset Selection}
\label{sec:datasets}

As aforementioned, existing studies tend to employ different datasets to evaluate the performance of the proposed method. In this study, we aim to provide a comprehensive and uniform comparison among existing techniques. Specifically, we selected the datasets with regard to the following criteria.

\begin{description}[leftmargin=*]
    \item[Criterion 1] \textbf{(Different Image Types):} We selected the datasets that associate to different image types (i.e., facial and non-facial), involve different sensitive attributes (e.g., age, race and gender), and are used for different classification tasks.

    \item[Criterion 2] \textbf{(Universal and Adaptable):} To make the adaptation of those studied methods easy to our experiment, we preferred the datasets that were widely used by existing studies.

\end{description}


As a result, we selected three datasets (i.e., CelebA, UTKFace, and CIFAR-10S) for conducting the empirical study. The summary of the datasets is shown in Table~\ref{tab:data}, which presents the sensitive attribute considered in the study, the labels of the classification tasks, and the studied methods that previously used the same settings. In addition, we also present the number of instances for model training and testing per each dataset by following existing studies.

\begin{table}[h]
    \caption{Summary of adopted benchmark datasets.}
    \label{tab:data}
    \centering
    \resizebox{0.98\columnwidth}{!}{%
        \begin{tabular}{l|ccc|lll}
        \toprule
            Dataset & Sensi. Attr. & Task Label & Used By & \#Train& \#Valid & \#Test\\
        \midrule
            CelebA & Gender & BlondHair & All& 93,141 & 19,867 & 19,962\\
            \hline
            \multirow{2}{*}{UTKFace}
            & Age & \multirow{2}{*}{Gender} & UW,BM,FLAC & 10,521 & 2,220 & 2,221\\
            & Race &  & UW,BM,BC+BB,MFD,FLAC& 10,744 & 2,370 & 2,370\\
            \hline
            CIFAR-10S & Color & Objects & OS,UW,BM,Adv,DI,MFD, & 50,000 & 2,000 & 18,000\\
        \bottomrule
        \end{tabular}
    }
\end{table}

Specifically, CelebA and UTKFace datasets represent human-centric datasets that involve sensitive attributes (e.g., age and race) reflected in facial images, while CIFAR-10S is a more general dataset in fairness study, which introduces biases related to image colors of objects (non-facial images). They were previously used for evaluation by partial/all methods under different conditions, e.g., using different metrics. From the table we can also find that previous studies tend to employ different datasets, hindering their comprehensive comparison. In this study, we aim to provide an extensive study through a systematic evaluation by employing all of them, which vary in multiple perspectives.


\subsection{Measurement Metrics}
\label{sec:metrics}

\subsubsection{Fairness Metrics}
\label{sec:fair_metrics}

Numerous prior studies have predominantly assessed fairness using a restricted set of fairness metrics. Although certain approaches have introduced their unique fairness metrics, these metrics have not achieved widespread adoption and lack broad applicability. To address these limitations, we employ the widely-used five metrics while excluding the self-defined metrics for measuring the fairness of models, i.e., \textbf{SPD}~\cite{DBLP:journals/csur/MehrabiMSLG21,DBLP:conf/sigsoft/BiswasR20,DBLP:conf/sigsoft/BiswasR21,DBLP:conf/sigsoft/ChenZSH22}, \textbf{DEO}~\cite{DBLP:journals/csur/MehrabiMSLG21,DBLP:conf/cvpr/JungLPM21,DBLP:conf/nips/ZhangZ0F00C22}, \textbf{EOD}~\cite{DBLP:journals/csur/MehrabiMSLG21,DBLP:conf/sigsoft/BiswasR20,DBLP:conf/sigsoft/BiswasR21,DBLP:conf/sigsoft/ChenZSH22}, \textbf{AAOD}~\cite{DBLP:conf/sigsoft/BiswasR20,DBLP:conf/sigsoft/BiswasR21,DBLP:conf/sigsoft/ChenZSH22}, and \textbf{AED}~\cite{DBLP:conf/sigsoft/BiswasR20,DBLP:conf/sigsoft/BiswasR21,DBLP:journals/corr/abs-2304-03935}, following previous work. The reason we chose these five metrics is twofold: (1) They have been integrated into the well-known \textit{AIF360} and \textit{Fairlearn} toolkits, indicating the importance and representativeness of them. (2) They have been well recognized and adopted by at least three previous studies shown in Table ~\ref{tab:works}.

Before delving into a detailed explanation of each metric, we will provide some definitions for the symbols used. Formally, let $A$ be a sensitive (or protected) attribute, and $A=1$ represents the instance that belongs to the privileged group while $A=0$ to the unprivileged group. We use $Y$ and $\hat{Y}$ to denote the expected and actual prediction labels, respectively, with 1 as the favorable label and 0 as the unfavorable label. We use $P$ to represent the probability of model prediction. Then, the definitions of the above metrics are defined as follows.

\begin{itemize}[leftmargin=*]

    \item \textbf{SPD} (Statistical Parity Difference): the difference of probabilities that the (un)privileged
    groups receive favorable outcomes:
    \begin{equation}
    \label{eq:spd}\small
    SPD = P[\hat{Y}=1|A=0] - P[\hat{Y}=1|A=1].
    \end{equation}
    \item \textbf{DEO} (Equalized Odds Difference): the max of differences about true/false positive rates 
    for unprivileged and privileged groups:
    \begin{equation}
    \label{eq:deo}
    \small
    \begin{split}
    DEO = max \{|P[\hat{Y}=1|A=0,Y=0] &- P[\hat{Y}=1|A=1,Y=0]|, \\ |P[\hat{Y}=1|A=0,Y=1] &- P[\hat{Y}=1|A=1,Y=1]|\}.
    \end{split}
    \end{equation}
    \item \textbf{EOD} (Equal Opportunity Difference): the true positive rate difference between unprivileged and privileged groups:
    \begin{equation}
    \label{eq:eod}
    \small
    \begin{split}
    EOD = P[\hat{Y}=1|A=0,Y=1]-P[\hat{Y}=1|A=1,Y=1].
    \end{split}
    \end{equation}
    \item \textbf{AAOD} (Average Absolute Odds Difference): the average of absolute difference of true/false positive rates 
    between unprivileged and privileged groups:
    \begin{equation}
    \label{eq:AAOD}  
    \small
    \begin{split}
    AAOD = \frac{1}{2} (|P[\hat{Y}=1|A=0,Y=0]&-P[\hat{Y}=1|A=1,Y=0]|\\ +|P[\hat{Y}=1|A=0,Y=1]&-P[\hat{Y}=1|A=1,Y=1]|).
    \end{split}
    \end{equation}
    \item \textbf{AED} (Accuracy Equality Difference): the difference of model misclassification rates across difference sensitive groups:
    \begin{equation}
    \label{eq:aed}  
    \small
    AED = P[\hat{Y} \neq Y|A=0] - P[\hat{Y} \neq Y|A=1].
    \end{equation}
    
\end{itemize}

Considering that the classification task for CIFAR-10S involves multi-class classification, whereas the classic fairness metrics mentioned earlier are primarily designed for binary classification tasks, we employ the \textbf{"ovr (one-vs-rest)"} strategy, inspired by ~\cite{DBLP:conf/nips/ZhangZ0F00C22,DBLP:conf/cvpr/JungLPM21} and \textit{sklearn} to decompose the N-ary classification problem into N binary classification subproblems. In our experiments, we initiate the process by computing the bias score for each class, following the same procedure as described for binary classification. Specifically, we calculate the fairness metrics for each class against the rest and then average the results across different classes to derive the multi-class versions of these fairness metrics.

According to the definitions of fairness metrics, a smaller metric value indicates a fairer model. 0 denotes absolute fairness.

\subsubsection{Performance Metrics}
\label{sec:acc_metrics}

To evaluate the quality of a model's predictions, we use traditional classification metrics, including \textbf{Accuracy} and \textbf{Balanced Accuracy}, which are defined by formulas~\ref{eq:acc} and~\ref{eq:bacc}. In particular, the latter can mitigate the effect of imbalanced dataset to the final result. 
In the formulas, \textit{TP/TN} denotes the number of instances that are correctly classified as positive/negative, while \textit{FP/FN} denotes the number of instances incorrectly classified as positive/negative.
\begin{equation}\small
    \label{eq:acc}  
\textit{Accuracy} = (TP + TN)/(TP+FP+FN+TN)
    \end{equation}
\begin{equation}\small
    \label{eq:bacc}  
    \textit{Balanced-Accuracy} = \left( TP/(TP + FN) + TN/(TN + FP)\right )/2
    \end{equation}
These two metrics are applicable to both binary and multi-class problems. For all metrics, larger values indicate better performance. The values of performance metrics are
between 0 and 1.

\subsection{Implementation Details}
\label{sec:details}





For all experiments, we use ResNet-18 ~\cite{DBLP:conf/cvpr/HeZRS16} as the backbone architecture by following existing studies~\cite{DBLP:journals/corr/abs-2209-15605,DBLP:journals/corr/abs-2304-14252,DBLP:conf/cvpr/JungLPM21,DBLP:conf/cvpr/WangDXZCW022,DBLP:conf/nips/ZhangZ0F00C22}.
Moreover, we used the best default configurations of the studied techniques provided in the respective papers.
Specifically, we firstly preprocessed the selected three datasets to ensure uniform data partitioning for consistency. Then, for each studied method, we utilized the optimal default configuration hyper-parameters provided in their respective papers to ensure the reliability of the experiments. However, since some methods originally did not utilize all of the studied datasets used in this paper, we adhered to their principles and adapted them to the new datasets , fine-tuning them for optimal results. In summary, to effectively and fairly compare the experimental results of each method, we standardized various factors such as data partitioning, model backbone, experiment repetition, experimental environment and so on.
In particular, to mitigate the effect of randomness, we have also repeated all our experiments \textbf{10 times} to ensure the reliability and stability of the results, which took about \textbf{40 days} apart from the costs of tool implementation and configuration. 
It's worth noting that since some approaches depend on a well-trained model (i.e., post-processing) while some approaches train the models from scratch (e.g., pre-processing), we compare the performance of the final optimized models directly. For methods that depend on existing well-trained models, we first conducted a grid search process to obtain the best-performing models as their inputs. We have published all the implementation details in our open-source repository.

Both our performance and fairness metrics are implemented based on \textit{sklearn}~\cite{scikit-learn} and \textit{Fairlearn}~\cite{DBLP:journals/corr/abs-2303-16626} libraries. Particularly, we have implemented them in an easy-to-use framework for future studies. It is available in our open-source repository. 
All experiments are implemented with Python 3.8.18 and PyTorch 1.10.1, and executed on an Ubuntu 20.04.6 LTS with 2.90GHz Intel(R) Xeon(R) Gold 6326 CPUs and eight NVIDIA GeForce RTX 3090 GPUs.

\subsection{Research Questions}
\label{sec:rqs}

Our empirical study focuses on the following research questions.

\begin{itemize}[leftmargin=*]

    \item \textbf{RQ1 (Overall effectiveness of fairness improving methods):} \textit{How do different fairness improvement methods perform in terms of accuracy and fairness?} 

    \item \textbf{RQ2 (Influence of evaluation metrics):} \textit{How do different accuracy and fairness metrics affect the evaluation results of DL models?}

    \item \textbf{RQ3 (Influence of datasets):} \textit{How do different dataset settings (dataset characteristics, variations in sensitive attributes and so on) impact fairness improvements?}

    \item \textbf{RQ4 (Efficiency of fairness improving methods):} \textit{What are the time costs of different fairness improvement methods?}
    
\end{itemize}

\section{Evaluation Results}
\label{sec:results}


\subsection{RQ1: Overall Effectiveness}

This research question assesses the overall performance of studied methods in terms of both the fairness and performance metrics. Our analysis takes both fairness and accuracy into consideration since an excellent fairness improving method should achieve good fairness scores without excessively sacrificing the accuracy, striking a balance or "trade-off" between the two.
Its objective is to 
see whether the best-performing method exists, and thereby offering valuable choices and insights for software engineering researchers engaged in addressing fairness-related issues.

\begin{figure}[htb]
    \begin{center}
    \includegraphics[width=0.98\columnwidth]{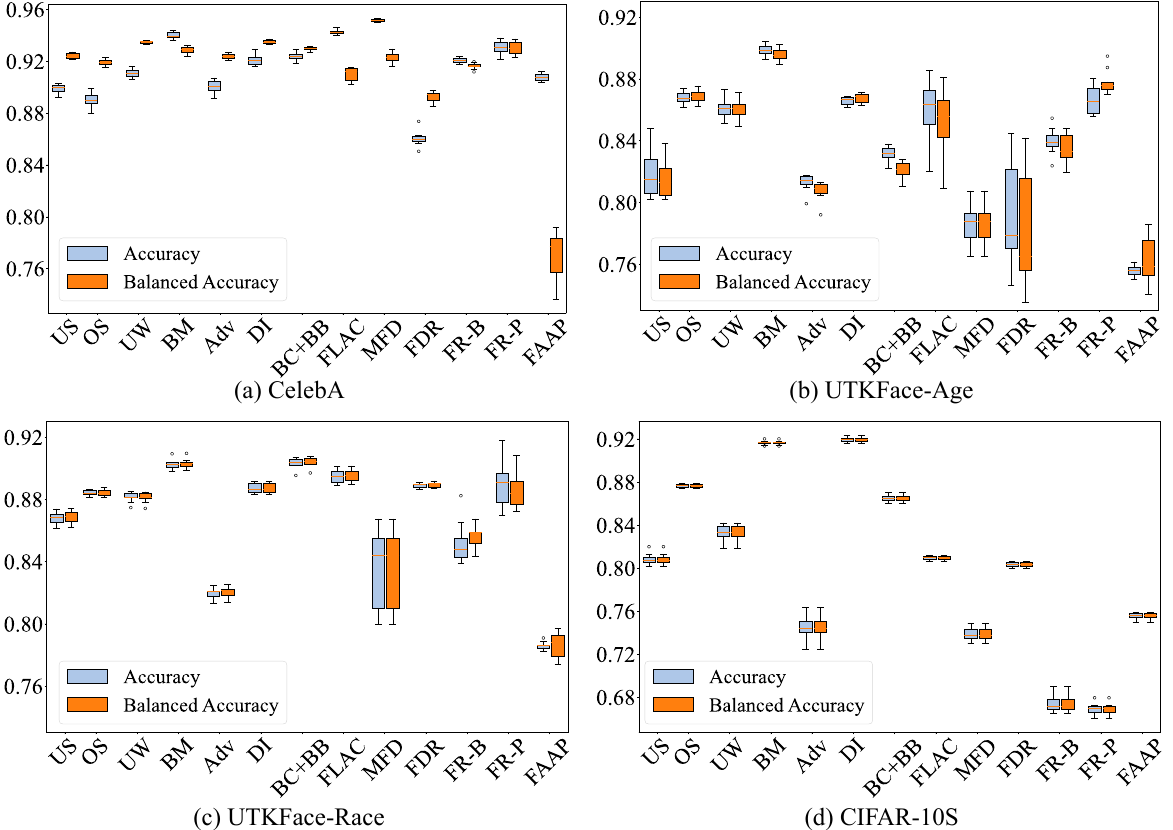}
    \end{center}
	\caption{Value distribution regarding accuracy metrics after applying different approaches on different datasets.}
	\label{fig:boxplot_acc}
\end{figure}

As introduced in Sections~\ref{sec:methods} and~\ref{sec:setup}, we have systematically studied 13 fairness improving methods over 3 diverse datasets.
The experimental results are presented in Table~\ref{tab:fair} and Figure~\ref{fig:boxplot_acc}. According to the experimental results, the performance of different approaches varies greatly over different datasets and even over different evaluation metrics. In particular, the best-performing method does not exist. For example, although the method DI performs better over the UTKFace-race and CIFAR-10S datasets than all the other methods, its performance over the other two datasets suffer a large decline compared with
many others, e.g., US and FDR. However, we can observe that the methods in pre-processing and in-processing categories 
are more effective to improve model fairness than those in the post-processing category, which yield less satisfactory results. 

Furthermore, pre-processing
and in-processing methods also tend to achieve higher model accuracy together with improving model fairness. This aligns
with our intuition, as the sources of bias in fairness mainly originate from the dataset and the
training procedure, making direct improvements in these aspects should be the most straightforward and effective approach.
Nevertheless, post-processing methods remain essential, particularly with the increasing prevalence of large-scale models in industry. When dealing with many already deployed industrial-scale
models, improving their fairness may only be achievable through fine-tuning specific layers or enhancing the entire model as a black box. Existing post-processing methods often employ adversarial
perturbations on input images or introduce fairness triggers to obscure the model's identification
of sensitive attributes so as to make fair predictions. However, we should acknowledge that their effectiveness still has much room for further improvement. More effective post-processing methods are urgent to be proposed.

\finding{While the best-performing fairness improvement method does not exist, pre-processing and in-processing methods have presented much better effectiveness than the post-processing methods.}

Considering the methods belonging the same categories, their effectiveness is very close although different methods may perform slightly better in some cases than the others for the pre-processing and in-processing methods regarding improving model fairness. For example, US slightly outperforms other pre-processing methods on the CelebA dataset, but slightly worse than BM on the CIFAR-10S dataset. Similarly, while FLAC and FDR achieve better model fairness than the other in-processing methods on the CelebA and UTKFace-age, they perform relatively poor than DI on the other two datasets. In contrast, FAAP in the post-processing category stands out as it achieve better than the other two methods (i.e., FR-B and FR-P) over almost all the datasets.
However, when taking the model accuracy into consideration, the conclusion will be different. For example, BM in the pre-processing category achieves the best accuracy over almost all the datasets. In fact, BM achieves the best model accuracy than almost all the other methods regardless of their categories. Recall the fairness performance of BM in Table~\ref{tab:fair}, we can reasonably conclude that BM achieves the best performance for balancing the model fairness and accuracy. 
It is important to note that BM actually combines pre-processing and in-processing techniques as it first samples the original dataset to create different distributions for each class and then it employs a specialized training and inference process to achieve the desired trade-off between accuracy and fairness by considering the distributions.
Furthermore, from Table~\ref{tab:fair} we can also observe that the performance of different methods over different datasets vary greatly, most of them tend to be affected. We leave a more comprehensive analysis on this point to Section~\ref{sec:result-rq3}.

\finding{BM achieves the best performance for balancing the model fairness and accuracy by combining the strength of both pre-processing and in-processing techniques.}


\begin{table*}[t]
    \caption{Result  comparison among different studied methods. The values in each cell denote the mean and standard deviation of the fairness metrics obtained from multiple experiments for the current method and dataset setting. We use different colors to highlight the values of different metrics, the darker of the color, the larger of the value, and the worse of the performance. 
    }
    \label{tab:fair}
    \centering
    \resizebox{\textwidth}{!}{
        \begin{tabular}{c|c|cccc|cccccc|ccc}
            \toprule
            \multirow{2}{*}{Dataset}                                                & \multirow{2}{*}{Metric} & \multicolumn{4}{c|}{Pre-processing}      & \multicolumn{6}{c|}{In-processing}       & \multicolumn{3}{c}{Post-processing}                                                                                                                                                                                                                                                                                                                                                                                                                                                    \\
                                                                                    &                         & US                                       & OS                                       & UW                                       & BM                                       & Adv                                      & DI                                       & BC+BB                                    & FLAC                                     & MFD                                      & FDR                                      & FR-B                                     & FR-P                                     & FAAP                                     \\
            \midrule
            \multirow{5}{*}{CelebA}                                                 & SPD                     & \cellcolor{red!19}$0.186_{\pm 0.004}$    & \cellcolor{red!34}$0.338_{\pm 0.007}$    & \cellcolor{red!26}$0.258_{\pm 0.004}$    & \cellcolor{red!21}$0.208_{\pm 0.003}$    & \cellcolor{red!33}$0.330_{\pm 0.007}$    & \cellcolor{red!23}$0.230_{\pm 0.006}$    & \cellcolor{red!27}$0.272_{\pm 0.006}$    & \cellcolor{red!17}$0.170_{\pm 0.004}$    & \cellcolor{red!21}$0.210_{\pm 0.003}$    & \cellcolor{red!14}$0.139_{\pm 0.010}$    & \cellcolor{red!29}$0.289_{\pm 0.004}$    & \cellcolor{red!27}$0.265_{\pm 0.005}$    & \cellcolor{red!19}$0.185_{\pm 0.005}$    \\
                                                                                    & DEO                     & \cellcolor{orange!05}$0.048_{\pm 0.007}$ & \cellcolor{orange!44}$0.438_{\pm 0.029}$ & \cellcolor{orange!18}$0.183_{\pm 0.023}$ & \cellcolor{orange!25}$0.247_{\pm 0.016}$ & \cellcolor{orange!50}$0.495_{\pm 0.027}$ & \cellcolor{orange!21}$0.207_{\pm 0.023}$ & \cellcolor{orange!40}$0.401_{\pm 0.025}$ & \cellcolor{orange!16}$0.160_{\pm 0.023}$ & \cellcolor{orange!40}$0.399_{\pm 0.026}$ & \cellcolor{orange!13}$0.127_{\pm 0.026}$ & \cellcolor{orange!58}$0.576_{\pm 0.011}$ & \cellcolor{orange!56}$0.559_{\pm 0.001}$ & \cellcolor{orange!52}$0.524_{\pm 0.002}$ \\
                                                                                    & EOD                     & \cellcolor{yellow!05}$0.047_{\pm 0.008}$ & \cellcolor{yellow!44}$0.438_{\pm 0.029}$ & \cellcolor{yellow!18}$0.183_{\pm 0.023}$ & \cellcolor{yellow!25}$0.247_{\pm 0.016}$ & \cellcolor{yellow!50}$0.495_{\pm 0.027}$ & \cellcolor{yellow!21}$0.207_{\pm 0.023}$ & \cellcolor{yellow!40}$0.401_{\pm 0.025}$ & \cellcolor{yellow!16}$0.160_{\pm 0.023}$ & \cellcolor{yellow!40}$0.399_{\pm 0.026}$ & \cellcolor{yellow!13}$0.127_{\pm 0.026}$ & \cellcolor{yellow!58}$0.576_{\pm 0.011}$ & \cellcolor{yellow!56}$0.559_{\pm 0.001}$ & \cellcolor{yellow!52}$0.524_{\pm 0.002}$ \\
                                                                                    & AAOD                    & \cellcolor{green!04}$0.041_{\pm 0.004}$  & \cellcolor{green!32}$0.315_{\pm 0.014}$  & \cellcolor{green!14}$0.144_{\pm 0.011}$  & \cellcolor{green!15}$0.149_{\pm 0.008}$  & \cellcolor{green!34}$0.337_{\pm 0.012}$  & \cellcolor{green!14}$0.141_{\pm 0.012}$  & \cellcolor{green!26}$0.258_{\pm 0.013}$  & \cellcolor{green!09}$0.090_{\pm 0.012}$  & \cellcolor{green!22}$0.224_{\pm 0.013}$  & \cellcolor{green!07}$0.068_{\pm 0.012}$  & \cellcolor{green!36}$0.358_{\pm 0.007}$  & \cellcolor{green!33}$0.327_{\pm 0.006}$  & \cellcolor{green!30}$0.299_{\pm 0.004}$  \\
                                                                                    & AED                     & \cellcolor{blue!02}$0.018_{\pm 0.003}$   & \cellcolor{blue!14}$0.142_{\pm 0.007}$   & \cellcolor{blue!08}$0.077_{\pm 0.004}$   & \cellcolor{blue!04}$0.042_{\pm 0.002}$   & \cellcolor{blue!13}$0.132_{\pm 0.007}$   & \cellcolor{blue!05}$0.053_{\pm 0.004}$   & \cellcolor{blue!09}$0.086_{\pm 0.004}$   & \cellcolor{blue!03}$0.028_{\pm 0.002}$   & \cellcolor{blue!04}$0.044_{\pm 0.002}$   & \cellcolor{blue!03}$0.026_{\pm 0.008}$   & \cellcolor{blue!10}$0.103_{\pm 0.001}$   & \cellcolor{blue!07}$0.073_{\pm 0.001}$   & \cellcolor{blue!11}$0.114_{\pm 0.001}$   \\ \hline
            \multirow{5}{*}{\begin{tabular}[c]{@{}c@{}}UTKFace\\ Age\end{tabular}}  & SPD                     & \cellcolor{red!11}$0.109_{\pm 0.018}$    & \cellcolor{red!46}$0.458_{\pm 0.016}$    & \cellcolor{red!23}$0.228_{\pm 0.037}$    & \cellcolor{red!10}$0.098_{\pm 0.036}$    & \cellcolor{red!60}$0.596_{\pm 0.011}$    & \cellcolor{red!49}$0.492_{\pm 0.022}$    & \cellcolor{red!15}$0.145_{\pm 0.020}$    & \cellcolor{red!08}$0.081_{\pm 0.035}$    & \cellcolor{red!05}$0.052_{\pm 0.029}$    & \cellcolor{red!13}$0.131_{\pm 0.071}$    & \cellcolor{red!52}$0.520_{\pm 0.004}$    & \cellcolor{red!50}$0.499_{\pm 0.004}$    & \cellcolor{red!48}$0.475_{\pm 0.005}$    \\
                                                                                    & DEO                     & \cellcolor{orange!26}$0.256_{\pm 0.027}$ & \cellcolor{orange!76}$0.755_{\pm 0.017}$ & \cellcolor{orange!43}$0.432_{\pm 0.050}$ & \cellcolor{orange!32}$0.315_{\pm 0.031}$ & \cellcolor{orange!86}$0.863_{\pm 0.015}$ & \cellcolor{orange!81}$0.810_{\pm 0.040}$ & \cellcolor{orange!27}$0.267_{\pm 0.032}$ & \cellcolor{orange!27}$0.269_{\pm 0.035}$ & \cellcolor{orange!27}$0.265_{\pm 0.031}$ & \cellcolor{orange!27}$0.269_{\pm 0.091}$ & \cellcolor{orange!81}$0.811_{\pm 0.009}$ & \cellcolor{orange!81}$0.806_{\pm 0.003}$ & \cellcolor{orange!73}$0.734_{\pm 0.003}$ \\
                                                                                    & EOD                     & \cellcolor{yellow!13}$0.127_{\pm 0.026}$ & \cellcolor{yellow!07}$0.073_{\pm 0.019}$ & \cellcolor{yellow!07}$0.069_{\pm 0.024}$ & \cellcolor{yellow!21}$0.213_{\pm 0.049}$ & \cellcolor{yellow!26}$0.256_{\pm 0.016}$ & \cellcolor{yellow!09}$0.086_{\pm 0.016}$ & \cellcolor{yellow!06}$0.056_{\pm 0.022}$ & \cellcolor{yellow!19}$0.190_{\pm 0.039}$ & \cellcolor{yellow!19}$0.185_{\pm 0.055}$ & \cellcolor{yellow!08}$0.075_{\pm 0.056}$ & \cellcolor{yellow!16}$0.155_{\pm 0.004}$ & \cellcolor{yellow!08}$0.075_{\pm 0.002}$ & \cellcolor{yellow!15}$0.153_{\pm 0.003}$ \\
                                                                                    & AAOD                    & \cellcolor{green!20}$0.196_{\pm 0.015}$  & \cellcolor{green!41}$0.414_{\pm 0.016}$  & \cellcolor{green!25}$0.251_{\pm 0.020}$  & \cellcolor{green!26}$0.263_{\pm 0.020}$  & \cellcolor{green!56}$0.560_{\pm 0.011}$  & \cellcolor{green!45}$0.448_{\pm 0.022}$  & \cellcolor{green!16}$0.161_{\pm 0.019}$  & \cellcolor{green!23}$0.228_{\pm 0.018}$  & \cellcolor{green!21}$0.213_{\pm 0.022}$  & \cellcolor{green!17}$0.171_{\pm 0.039}$  & \cellcolor{green!49}$0.486_{\pm 0.003}$  & \cellcolor{green!44}$0.435_{\pm 0.004}$  & \cellcolor{green!44}$0.442_{\pm 0.005}$  \\
                                                                                    & AED                     & \cellcolor{blue!20}$0.196_{\pm 0.016}$   & \cellcolor{blue!34}$0.335_{\pm 0.008}$   & \cellcolor{blue!25}$0.247_{\pm 0.019}$   & \cellcolor{blue!27}$0.266_{\pm 0.020}$   & \cellcolor{blue!31}$0.306_{\pm 0.010}$   & \cellcolor{blue!35}$0.354_{\pm 0.020}$   & \cellcolor{blue!18}$0.177_{\pm 0.018}$   & \cellcolor{blue!24}$0.239_{\pm 0.017}$   & \cellcolor{blue!21}$0.213_{\pm 0.022}$   & \cellcolor{blue!19}$0.185_{\pm 0.035}$   & \cellcolor{blue!33}$0.326_{\pm 0.003}$   & \cellcolor{blue!40}$0.404_{\pm 0.001}$   & \cellcolor{blue!28}$0.277_{\pm 0.001}$   \\ \hline
            \multirow{5}{*}{\begin{tabular}[c]{@{}c@{}}UTKFace\\ Race\end{tabular}} & SPD                     & \cellcolor{red!03}$0.031_{\pm 0.007}$    & \cellcolor{red!08}$0.079_{\pm 0.005}$    & \cellcolor{red!05}$0.045_{\pm 0.007}$    & \cellcolor{red!02}$0.023_{\pm 0.009}$    & \cellcolor{red!24}$0.240_{\pm 0.012}$    & \cellcolor{red!03}$0.029_{\pm 0.008}$    & \cellcolor{red!05}$0.052_{\pm 0.006}$    & \cellcolor{red!05}$0.045_{\pm 0.012}$    & \cellcolor{red!04}$0.036_{\pm 0.014}$    & \cellcolor{red!07}$0.066_{\pm 0.007}$    & \cellcolor{red!20}$0.200_{\pm 0.006}$    & \cellcolor{red!17}$0.168_{\pm 0.006}$    & \cellcolor{red!15}$0.153_{\pm 0.005}$    \\
                                                                                    & DEO                     & \cellcolor{orange!02}$0.021_{\pm 0.007}$ & \cellcolor{orange!07}$0.068_{\pm 0.009}$ & \cellcolor{orange!03}$0.029_{\pm 0.009}$ & \cellcolor{orange!02}$0.024_{\pm 0.010}$ & \cellcolor{orange!23}$0.234_{\pm 0.015}$ & \cellcolor{orange!02}$0.016_{\pm 0.007}$ & \cellcolor{orange!04}$0.040_{\pm 0.007}$ & \cellcolor{orange!04}$0.041_{\pm 0.019}$ & \cellcolor{orange!05}$0.049_{\pm 0.011}$ & \cellcolor{orange!07}$0.069_{\pm 0.012}$ & \cellcolor{orange!18}$0.179_{\pm 0.003}$ & \cellcolor{orange!15}$0.150_{\pm 0.002}$ & \cellcolor{orange!17}$0.166_{\pm 0.003}$ \\
                                                                                    & EOD                     & \cellcolor{yellow!01}$0.011_{\pm 0.009}$ & \cellcolor{yellow!06}$0.055_{\pm 0.015}$ & \cellcolor{yellow!02}$0.023_{\pm 0.012}$ & \cellcolor{yellow!02}$0.018_{\pm 0.010}$ & \cellcolor{yellow!23}$0.228_{\pm 0.021}$ & \cellcolor{yellow!01}$0.008_{\pm 0.006}$ & \cellcolor{yellow!02}$0.016_{\pm 0.005}$ & \cellcolor{yellow!01}$0.011_{\pm 0.007}$ & \cellcolor{yellow!04}$0.037_{\pm 0.016}$ & \cellcolor{yellow!02}$0.016_{\pm 0.007}$ & \cellcolor{yellow!18}$0.179_{\pm 0.003}$ & \cellcolor{yellow!13}$0.130_{\pm 0.002}$ & \cellcolor{yellow!09}$0.090_{\pm 0.003}$ \\
                                                                                    & AAOD                    & \cellcolor{green!01}$0.014_{\pm 0.005}$  & \cellcolor{green!06}$0.057_{\pm 0.005}$  & \cellcolor{green!02}$0.023_{\pm 0.007}$  & \cellcolor{green!02}$0.016_{\pm 0.009}$  & \cellcolor{green!22}$0.222_{\pm 0.012}$  & \cellcolor{green!01}$0.010_{\pm 0.005}$  & \cellcolor{green!03}$0.028_{\pm 0.006}$  & \cellcolor{green!03}$0.026_{\pm 0.010}$  & \cellcolor{green!04}$0.038_{\pm 0.011}$  & \cellcolor{green!04}$0.042_{\pm 0.007}$  & \cellcolor{green!18}$0.177_{\pm 0.004}$  & \cellcolor{green!14}$0.139_{\pm 0.003}$  & \cellcolor{green!13}$0.128_{\pm 0.006}$  \\
                                                                                    & AED                     & \cellcolor{blue!01}$0.010_{\pm 0.006}$   & \cellcolor{blue!01}$0.012_{\pm 0.006}$   & \cellcolor{blue!01}$0.006_{\pm 0.006}$   & \cellcolor{blue!02}$0.015_{\pm 0.009}$   & \cellcolor{blue!01}$0.009_{\pm 0.008}$   & \cellcolor{blue!01}$0.007_{\pm 0.004}$   & \cellcolor{blue!01}$0.012_{\pm 0.003}$   & \cellcolor{blue!02}$0.019_{\pm 0.011}$   & \cellcolor{blue!01}$0.014_{\pm 0.007}$   & \cellcolor{blue!03}$0.028_{\pm 0.007}$   & \cellcolor{blue!01}$0.007_{\pm 0.001}$   & \cellcolor{blue!02}$0.016_{\pm 0.001}$   & \cellcolor{blue!05}$0.048_{\pm 0.002}$   \\ \hline
            \multirow{5}{*}{CIFAR-10S}                                              & SPD                     & \cellcolor{red!01}$0.003_{\pm 0.001}$    & \cellcolor{red!03}$0.026_{\pm 0.001}$    & \cellcolor{red!03}$0.033_{\pm 0.002}$    & \cellcolor{red!01}$0.002_{\pm 0.001}$    & \cellcolor{red!07}$0.069_{\pm 0.004}$    & \cellcolor{red!01}$0.002_{\pm 0.001}$    & \cellcolor{red!03}$0.032_{\pm 0.001}$    & \cellcolor{red!03}$0.029_{\pm 0.001}$    & \cellcolor{red!05}$0.045_{\pm 0.017}$    & \cellcolor{red!02}$0.016_{\pm 0.002}$    & \cellcolor{red!07}$0.069_{\pm 0.001}$    & \cellcolor{red!07}$0.069_{\pm 0.001}$    & \cellcolor{red!02}$0.015_{\pm 0.005}$    \\
                                                                                    & DEO                     & \cellcolor{orange!02}$0.015_{\pm 0.001}$ & \cellcolor{orange!11}$0.108_{\pm 0.002}$ & \cellcolor{orange!13}$0.131_{\pm 0.009}$ & \cellcolor{orange!01}$0.012_{\pm 0.001}$ & \cellcolor{orange!28}$0.282_{\pm 0.019}$ & \cellcolor{orange!01}$0.010_{\pm 0.001}$ & \cellcolor{orange!13}$0.128_{\pm 0.006}$ & \cellcolor{orange!12}$0.120_{\pm 0.004}$ & \cellcolor{orange!17}$0.174_{\pm 0.064}$ & \cellcolor{orange!07}$0.074_{\pm 0.008}$ & \cellcolor{orange!25}$0.249_{\pm 0.002}$ & \cellcolor{orange!25}$0.250_{\pm 0.004}$ & \cellcolor{orange!08}$0.082_{\pm 0.003}$ \\
                                                                                    & EOD                     & \cellcolor{yellow!02}$0.015_{\pm 0.001}$ & \cellcolor{yellow!11}$0.108_{\pm 0.002}$ & \cellcolor{yellow!13}$0.131_{\pm 0.009}$ & \cellcolor{yellow!01}$0.012_{\pm 0.001}$ & \cellcolor{yellow!28}$0.282_{\pm 0.019}$ & \cellcolor{yellow!01}$0.010_{\pm 0.001}$ & \cellcolor{yellow!13}$0.128_{\pm 0.006}$ & \cellcolor{yellow!12}$0.120_{\pm 0.004}$ & \cellcolor{yellow!17}$0.174_{\pm 0.064}$ & \cellcolor{yellow!07}$0.074_{\pm 0.008}$ & \cellcolor{yellow!25}$0.249_{\pm 0.002}$ & \cellcolor{yellow!25}$0.250_{\pm 0.004}$ & \cellcolor{yellow!08}$0.082_{\pm 0.003}$ \\
                                                                                    & AAOD                    & \cellcolor{green!01}$0.008_{\pm 0.001}$  & \cellcolor{green!06}$0.063_{\pm 0.001}$  & \cellcolor{green!08}$0.076_{\pm 0.005}$  & \cellcolor{green!01}$0.007_{\pm 0.001}$  & \cellcolor{green!16}$0.163_{\pm 0.011}$  & \cellcolor{green!01}$0.006_{\pm 0.001}$  & \cellcolor{green!07}$0.074_{\pm 0.004}$  & \cellcolor{green!07}$0.069_{\pm 0.003}$  & \cellcolor{green!10}$0.102_{\pm 0.038}$  & \cellcolor{green!04}$0.042_{\pm 0.005}$  & \cellcolor{green!15}$0.149_{\pm 0.003}$  & \cellcolor{green!15}$0.150_{\pm 0.003}$  & \cellcolor{green!05}$0.048_{\pm 0.003}$  \\
                                                                                    & AED                     & \cellcolor{blue!01}$0.002_{\pm 0.001}$   & \cellcolor{blue!01}$0.005_{\pm 0.001}$   & \cellcolor{blue!01}$0.007_{\pm 0.001}$   & \cellcolor{blue!01}$0.001_{\pm 0.001}$   & \cellcolor{blue!02}$0.015_{\pm 0.001}$   & \cellcolor{blue!01}$0.001_{\pm 0.001}$   & \cellcolor{blue!01}$0.006_{\pm 0.001}$   & \cellcolor{blue!01}$0.006_{\pm 0.001}$   & \cellcolor{blue!01}$0.011_{\pm 0.005}$   & \cellcolor{blue!01}$0.004_{\pm 0.001}$   & \cellcolor{blue!02}$0.019_{\pm 0.001}$   & \cellcolor{blue!02}$0.019_{\pm 0.001}$   & \cellcolor{blue!01}$0.005_{\pm 0.001}$   \\
            \bottomrule
        \end{tabular}
    }
\end{table*}

For further analysis of the reasons for the different effectiveness of methods belonging to the same category, we first take the pre-processing category as an example. We believe that the main reason is due to their underlying principles and their suitability for addressing class imbalance and bias in the data. For example, US works well when the class imbalance is significant, as it directly addresses this issue by reducing the majority class. Additionally, by selectively dropping samples from specific subgroups, it can effectively mitigate bias by balancing the sizes of these subgroups. However, OS may result in over-fitting, especially with complex models like deep neural networks, when samples are excessively duplicated. This can lead to poorer generalization on the test set, which might explain its lower performance. To our surprise, as a newly proposed method, BM might not always outperform other techniques, such as US, in terms of fairness. The possible reason is that BM might introduce additional complexity to the pre-processing pipeline by emulating bias distributions across classes. This added complexity may lead to over-fitting or decreased generalization performance. In this case, a simpler method like US might suffice and even outperform BM on fairness. However, as aforementioned, when considering both model fairness and accuracy, BM should still be the best.

\finding{For pre-processing methods, over-fitting often proves to be a key factor affecting the performance of various methods.}

Similarly, for in-processing methods, we observe that although we cannot determine which method performs the best, the Adv method exhibits the poorest fairness performance across all datasets, especially on the UTKFace-age as shown in Table~\ref{tab:fair}. The core idea behind Adv is \textit{fairness through blindness}~\cite{DBLP:conf/cvpr/WangQKGNHR20}, which means attempting to prevent a model from explicitly encoding information about sensitive attributes through adversarial training. However, as we have observed, this practice of ignoring sensitive attribute encoding has several issues. Firstly, due to the phenomenon of redundant encoding~\cite{DBLP:conf/nips/HardtPNS16,DBLP:conf/innovations/DworkHPRZ12}, even if there is no particular sensitive attribute in the classifier's feature representation, combinations of other attributes can be used as a proxy. For instance, considering a real-world task where a bank assesses a loan application without considering the applicant's gender. If the applicant's occupation is \textit{nurse}, we can still infer with a high probability that the applicant is likely female based on the \textit{nurse} attribute. Secondly, this practice of discarding sensitive attribute encoding may harm accuracy, and the results in Figure~\ref{fig:boxplot_acc} support this hypothesis, with Adv's lowest accuracy across all datasets among in-processing methods. The reason may be that internal connections between different attributes likely have a significant impact on accuracy.

\finding{Sensitive attributes play an important role in model predictions. Simply improving fairness by removing the encoding of sensitive attributes may not be an effective approach.}

Finally, for post-processing methods, we observe that all the studied methods have shown unstable performance. The reason may be that these methods share a common optimization objective, which is the min-max adversarial training framework. In this framework, the discriminator attempts to improve its predictions of sensitive attributes, while the fairness trigger (in FR-B and FR-P) or the generator (in FAAP) modifies input images to reduce the information about protected attributes in the latent representation of a deployed black-box model, thus attempting to degrade predictions. These methods primarily aim to mitigate model bias by hiding the information of sensitive attributes during the feature extraction process, so that the model does not associate predictions with sensitive attributes. However, this approach has its drawbacks. Hiding information about sensitive attributes may also hide features closely related to those attributes, ultimately affecting both accuracy and fairness performance. 
Supposing an AI judger makes predictions based on human faces, 
if we attempt to hide racial encoding information in facial images, we may unintentionally affect information related to skin color, which is closely linked to race. The loss of this information can potentially impact the performance of other prediction tasks.

\finding{For post-processing methods, hiding sensitive information encoding may not be an ideal choice, and more effective approaches are still in urgent need.}



\subsection{RQ2: Influence of evaluation metrics}

In this RQ, we investigate the performance of existing methods over different metrics, aiming to inspire researchers to choose appropriate metrics for evaluating DL models in future studies.

    

\begin{figure}[htb]
    \centering
    \includegraphics[width=\columnwidth]{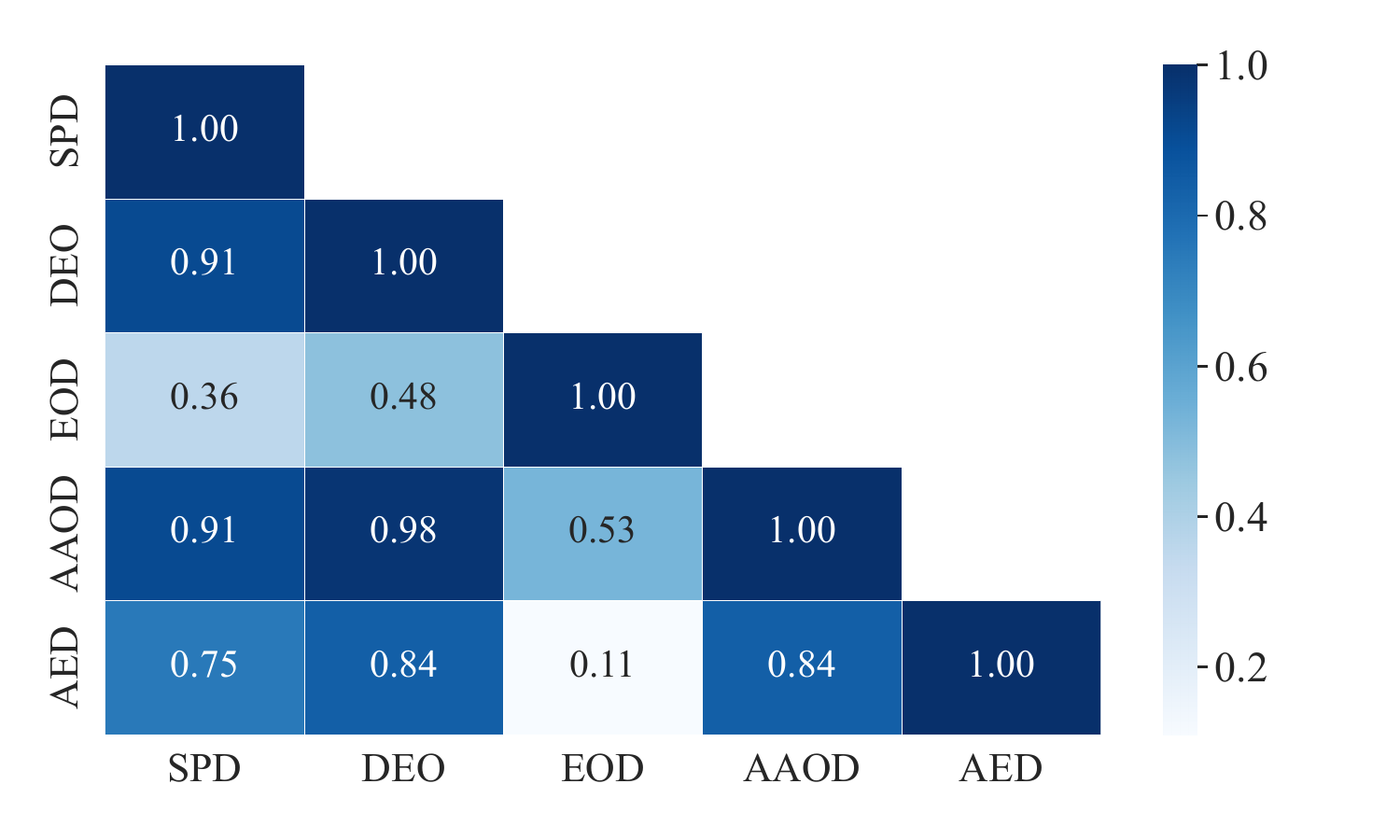}
    \caption{Pearson correlation across different metrics.} 
	\label{fig:corr}
\end{figure}

Regarding fairness metrics, we find that if a method performs well on one metric, it tends to perform well on other metrics as well, and vice versa. From Table~\ref{tab:fair}, we can also observe that the value distributions of each metric are highly similar within each dataset. Although different fairness metrics evaluate the performance of DL models from different aspects, when designing fairness improving methods, common strategies or techniques are often employed, including using fairness metrics as optimization objectives and balancing sampling, to ensure good performance across different metrics. 
To more clearly understand the relations between different fairness metrics, we have calculated the Pearson Correlation coefficient~\cite{enwiki:1177721736} across different metrics over all datasets. The result shown in Figure~\ref{fig:corr} further confirms our hypothesis that all fairness metrics are positively correlated with each other.


\finding{Existing fairness improvement approaches tend to be \textit{insensitive} to the fairness metrics.
That is, if a method performs well regarding one metric, it also tends to perform well regarding the other metrics.}

Moreover, we conducted a further in-depth investigation into the correlations between different fairness metrics and aimed to provide valuable guidance for future studies. From Figure~\ref{fig:corr} we can observe that AAOD has the highest overall correlation as its correlation coefficients with the other four metrics never fall below 0.5. We analyze the reasons based on the definitions of each fairness metric (see Section ~\ref{sec:fair_metrics}). The goal of SPD is to ensure a DL model’s predictions are independent of membership in a sensitive group. It is an easy metric but it only considers a single binary protected attribute, which is sometimes insufficient. For DEO, EOD and AAOD, the goal of these metrics is to ensure a DL model performs equally well for different groups. These metrics are stricter than SPD because they require consideration of both the sensitive attribute and the true labels, involving the measurement of the difference between the true positive rate and false positive rate for the \textit{unprivileged} and \textit{privileged} groups. 
AAOD, as an arithmetic average of the difference in false positive rates and the difference in true positive rates, effectively balances the effects of both difference metrics. This is supported by the experimental results, where AAOD generally exhibits higher differentiation compared to DEO and EOD. Therefore, AAOD appears to be a robust fairness evaluation metric compared to DEO and EOD, especially in scenarios where DEO and EOD show substantial discrepancies. AED measures the difference in misclassification rates of the model on different sensitive attribute groups. The major drawback of this metric is that its values are generally small, particularly on the CIFAR-10S dataset, resulting in lower differentiability compared to SPD and AAOD. Therefore, it is not considered an excellent fairness evaluation metric.


\finding{
AAOD emerges as the most promising and representative metric for evaluating model fairness.}

When considering the two performance metrics, i.e., \textit{accuracy} and \textit{balanced accuracy}, they are very close to each other according to the results shown in Figure~\ref{fig:boxplot_acc}. In particular, on UTKFace and CIFAR-10S, accuracy and balanced accuracy results are almost equal. This indicates that the distribution of sensitive attributes on these datasets is balanced, in which case the two metrics are equivalent. However, on the CelebA dataset, we find that there is a significant difference between these two accuracy metrics, especially evident in the case of the FAAP method. 
It suggests that the FAAP method is highly sensitive to imbalanced class distributions in CelebA. We have mentioned earlier that FAAP, as a post-processing method, heavily relies on modifying input data to improve fairness. However, the imbalanced data distribution may make the model more biased towards the majority class, thereby affecting its normal predictions. After class balancing, we observe a significant decrease in accuracy. From Figure ~\ref{fig:boxplot_acc}, we also find that while FAAP exhibits substantial differences, there are still many methods that show consistent accuracy on the CelebA dataset, such as BM, BC+BB, and FR-P. Nevertheless, it is better to consider the two metrics together for a systematic and comprehensive analysis of the results.

\finding{For an imbalanced dataset, accuracy and balanced accuracy tend to produce diverse results, and thus both of them should be used to comprehensively measure the performance of fairness improvement methods.
}

\subsection{RQ3: Influence of datasets}
\label{sec:result-rq3}

Deep learning models tend to yield varying results on different datasets. An excellent fairness improvement method should perform competitively on different datasets. However, existing studies only use limited datasets. Therefore, this RQ aims to comprehensively analyze the performance of different fairness improving methods on various datasets.

For each method, we find that both accuracy and fairness vary significantly across different datasets according to the results shown in Table~\ref{tab:fair} and Figures~\ref{fig:boxplot_acc}. Regarding fairness, the results from Table~\ref{tab:fair} indicate that the overall fairness scores on CelebA and UTKFace-age are darker than the other two datasets. This suggests that for each method, fairness performance on CelebA and UTKFace-age is inferior to that on the other datasets. On the other hand, when considering accuracy, we have also observed that methods on the CelebA dataset have significantly higher accuracy compared to the other datasets, while CIFAR-10S exhibits the lowest overall accuracy. Particularly, we notice that the FR-B and FR-P methods, which have decent accuracy on other datasets, perform catastrophically poorly with accuracy below 0.70 on CIFAR-10S. Our analysis suggests that this should be due to the overreliance of FR methods on input data. CIFAR-10S, being an artificial dataset, has sensitive attributes that are artificially constructed (color or grayscale), and FR methods may not make appropriate modifications to such sensitive attributes during image alteration, thus disrupting the model's normal accuracy. Similarly, we find that what may be a superior method on one dataset doesn't necessarily hold when applied to a different dataset. For instance, by examining the table, we observe that on UTKFace-age, BC+BB exhibits significantly better fairness than DI, but conversely, on the other three datasets, DI shows substantial advantages. This indicates that existing methods are sensitive to the dataset they are applied to, and there is no best-performing method on all datasets.

\finding{Existing methods are \textit{sensitive} to different datasets in the evaluation, which are supported by two observations: 1) For a certain method, it tends to achieve different fairness and accuracy over different datasets; 2) For two different methods, one may outperform the other on certain dataset A, but underperform the other on another dataset B. Therefore, the conclusions from one certain dataset are hard to generalize to others.}

There is another point worth our attention. In the UTKFace dataset, age and race, two sensitive attributes, yield markedly different results. When age is treated as the sensitive attribute, the overall accuracy and fairness of various methods are lower compared to when race is the sensitive attribute. This difference is particularly pronounced in terms of fairness. By observing the table, we can see that, even within the UTKFace dataset, the colors for age are noticeably darker than those for race, indicating the differences in fairness across various methods due to different sensitive attributes.
Therefore, the choice of sensitive attributes is crucial in fairness tasks since different methods are suited for different ones.

\subsection{RQ4: Efficiency of studied methods}

The first three RQs explore the impact of different factors.
In practical applications, the time efficiency of each method is also an important consideration for method selection.
Therefore, our final RQ primarily investigates the time cost of different methods. Since the optimal experimental hyperparameters vary for different methods (such as batch size and epochs), we decide to measure the efficiency of each method by averaging the time cost for each epoch. It's important to note that all our experimental results are the averages obtained after conducting multiple experiments.

Table ~\ref{tab:time} presents the time cost per epoch of each fairness improving method. In the table, we highlight the highest efficiency in \textcolor{gray}{gray} color regarding the complete training time, while {\ul{underline}} the shortest time regarding a single epoch. 
Please note that analyzing both types (complete \textit{vs} single) of time cost have their values. The complete training time reflects the overall complexity of methods under their optimal parameter settings, while the time per epoch indicates the intrinsic computation complexity of the optimizing process of the method’s effectiveness in practical use. 

From the table, we can observe that FDR is much more efficient than other approaches considering a single training epoch. The reason is that FDR improves model fairness only through last-layer fine-tuning by incorporating fairness constraints and data reweighting ~\cite{DBLP:journals/corr/abs-2304-03935}. These operations only require updating the parameters of the last layer, resulting in very short runtime and high efficiency. In comparison, other in-processing methods incur significantly higher time overheads. This is primarily because these methods either introduce new loss functions and optimization objectives, design novel model modules, or employ techniques like distillation, all of which involve learning a large number of parameters during the training process, making it time-consuming. In contrast, considering the complete training time, there is no individual method outperforms all the others. For example,
OS is the most efficient on UTKFace while FAAP is the most efficient on CIFAR-10S. As previously analyzed, BM emerges as the most effective method, but it is not the most efficient one over every studied datasets. Similarly, FDR, which exhibits overwhelming efficiency and good fairness improvement among the in-processing methods, comes at the cost of significant accuracy sacrifices. Therefore, end users should to choose the most suitable method based on the specific task’s varying emphasis on effectiveness and efficiency.

\begin{table}[t]
    \caption{Average time cost per epoch of each method for improving model fairness (in seconds).}
    \label{tab:time}
    \centering
    \resizebox{0.98\columnwidth}{!}{
    \begin{tabular}{c|c|rr|rr|rr|rr}
        \toprule
        \multirow{2}{*}{Category}   & \multirow{2}{*}{Method} & \multicolumn{2}{c|}{CelebA} & \multicolumn{2}{c|}{UTKFace Age} & \multicolumn{2}{c|}{UTKFace Race} & \multicolumn{2}{c}{CIFAR-10S} \\
                                    &                         & t/epoch       & epochs      & t/epoch         & epochs         & t/epoch          & epochs         & t/epoch        & epochs       \\
        \midrule
        \multirow{4}{*}{Pre-proc.}  & US                      & 81.4          & 170         & 11.5            & 400            & 15.6             & 120            & 14.8           & 2000         \\
                                    & OS                      & 590.3         & 4           & \hl{16.6}       & \hl{7}              & \hl{13.7}        & \hl{10}             & 65.0           & 100          \\
                                    & UW                      & 252.9         & 10          & 8.5             & 20             & 8.9              & 20             & 71.5           & 200          \\
                                    & BM                      & 528.2         & 10          & 14.3            & 20             & 15.5             & 20             & 72.2           & 200          \\ \hline
        \multirow{6}{*}{In-proc.}   & Adv                     & 242.5         & 10          & 13.3            & 20             & 14.1             & 20             & 72.6           & 200          \\
                                    & DI                      & 242.3         & 10          & 13.3            & 20             & 13.5             & 20             & 68.1           & 200          \\
                                    & BC+BB                   & 564.3         & 10          & 23.3            & 20             & 23.5             & 20             & 67.9           & 200          \\
                                    & FLAC                    & 277.2         & 10          & 14.5            & 20             & 13.4             & 20             & 40.5           & 200          \\
                                    & MFD                     & 523.8         & 50          & 70.6            & 100            & 48.7             & 100            & 106.4          & 50           \\
                                    & FDR                     & \hl{\ul{2.2}} & \hl{1000}        & \ul{0.6}        & 1500           & \ul{0.8}         & 1500           & \ul{1.7}       & 1000         \\ \hline
        \multirow{3}{*}{Post-proc.} & FR-B                    & 379.8         & 20          & 41.8            & 20             & 42.7             & 20             & 76.7           & 20           \\
                                    & FR-P                    & 586.3         & 20          & 46.8            & 20             & 43.4             & 20             & 74.0           & 20           \\
                                    & FAAP                    & 288.1         & 50          & 17.1            & 50             & 17.5             & 50             & \hl{14.5}      & \hl{50}           \\
        \bottomrule
    \end{tabular}
    }
\end{table}

\finding{An excellent fairness improving method should strike a balance between effectiveness and efficiency, ensuring that it not only improves fairness but also maintains acceptable levels of accuracy without consuming excessive computational resources.}


\section{Implications}
\label{sec:implication}

Based on our result analysis, we have summarized several implications that can facilitate future research in this research area.

\textbf{Combining the strength of different methods}. According to our empirical results (i.e. Finding 1), there is no best-performing methods in all application scenarios and different methods are conceptually complementary. For example, pre-processing methods focus more on transforming or augmenting training data, while in-processing methods primarily optimize the design of objective functions and model structures. Our experimental results also confirm that. Therefore, it should be promising to combine the strength of individual methods since they improve the models from different perspectives, e.g., training data and training process. In particular, directly combining the results of different methods can be effective. However, investigating the deep combination of different methods by considering their core novelty still needs more exploration.

\textbf{Developing more effective post-processing methods}. With the rapid development and inspiring performance of large language models (LLMs), they have been adopted in various applications. However, re-training LLMs can be a hard or even impossible task for end users due to the high requirement on computing resources and the reliance on training data. Besides, for deployed models, post-processing methods are intuitively the best choice for improving model fairness. Nevertheless, according to Finding 5, the results shown in Table~\ref{tab:fair} demonstrate that existing post-processing methods are not satisfactory and still have much room for further improvement compared with the other two categories. Therefore, more studies should focus on the post-processing methods.

\textbf{Understanding the source of unfairness}. Based on the introduction of existing approaches, almost all of them aim to improve the model fairness in somehow a blind way without understanding the core reasons of the unfairness in DL models. Specifically, pre-processing methods assume that the training data contain bias while in-processing methods deem the bias is involved during the training process. However, even adopting the same training data and the same training process (see Section~\ref{sec:result-rq3}), the model performance regarding different attributes can also be different. Therefore, understanding the source of unfairness in the DL models may help to better overcome this issue. In particular, model interpretation techniques~\cite{DBLP:conf/ijcai/0002WHZFZZ21,DBLP:conf/iccv/ChenCHRZ19} can be incorporated to facilitate this task.


\textbf{Incorporating the correlations among different attributes deeply.} Indeed, sensitive attributes play a crucial role in model predictions regarding result fairness. However, according to Finding 4, simply removing the sensitive attributes does not necessarily improve model fairness, and incorporating the correlations between the sensitive attributes and the others is also important. In fact, the existing study~\cite{DBLP:conf/icse/LiMC0WZX22} has already considered such information for improving model fairness. Nevertheless, it was designed for the structured (i.e., tabular) data, which should be much simpler than the images since the image features are not clearly defined.
In other words, how to effectively identify the implicit correlations between different image features and incorporate them into the process of fairness improvement still requires more in-depth exploration. 

\section{Threats to Validity}
\label{sec:theats}

The \textit{external} threats to validity mainly lie in the data selection in our study. In order to comprehensively evaluate the performance of existing methods, we have employed three widely-used datasets involving different sensitive attributes and classification tasks. In addition, according to our findings that different methods tend to have different effectiveness. Although our results cannot generalize to other datasets, the findings should not be affected.

The \textit{internal} threat to validity mainly lies in the implementations of the studies methods. To mitigate this threat, we mainly employed the open-source implementations of the corresponding papers if available, while for other methods, we have double-checked their performance with the corresponding authors to ensure they are correctly implemented and configured. Furthermore, to ease the replication and promote future research, we have made all our experimental results and implementations open-source.



\section{Related Work}
\label{sec:related}


\subsection{Fairness Improving Methods}
\textit{Pre-processing Methods}
Pre-processing methods ~\cite{DBLP:conf/cvpr/RamaswamyKR21,DBLP:journals/csur/MehrabiMSLG21,DBLP:journals/corr/abs-2211-00168,DBLP:conf/cvpr/QuadriantoST19,DBLP:conf/mm/ZhangS20} aim to remove underlying discrimination by calibrating training data to eliminate spurious correlations and training fairer models on the modified data. 
Ramaswamy \textit{et al.} ~\cite{DBLP:conf/cvpr/RamaswamyKR21} proposed a GAN-based data augmentation method to  balance the  training data. 
Yao \textit{et al.} ~\cite{DBLP:journals/corr/abs-2211-00168} proposed the 
methods 
to maintain useful information while filtering out bias information. 
Quadrianto \textit{et al.} ~\cite{DBLP:conf/cvpr/QuadriantoST19} learned a mapping from an input domain to a fair target domain to mitigate gender bias. 
Zhang \textit{et al.} ~\cite{DBLP:conf/mm/ZhangS20} employed adversarial examples to balance the training data for visual debiasing. 

\textit{In-processing Methods}
In-processing techniques ~\cite{DBLP:journals/csur/MehrabiMSLG21,DBLP:conf/eccv/LokhandeARS20,DBLP:conf/eccv/KehrenbergBTQ20,DBLP:conf/iclr/Roh0WS21,DBLP:conf/cvpr/XuHSLLHL021,DBLP:conf/eccv/LinKJ22,DBLP:conf/cvpr/ParkLLHKB22} focus on modifying and changing learning algorithms to remove discrimination during training. 
Lokhande \textit{et al.} ~\cite{DBLP:conf/eccv/LokhandeARS20} offered a simplified approach by treating fairness measures as constraints on the model's output, which is incorporated through an augmented Lagrangian framework. 
Kehrenberg \textit{et al.} ~\cite{DBLP:conf/eccv/KehrenbergBTQ20} proposed NIFR 
model to learn invariant representations to improve algorithmic fairness.
Roh \textit{et al.} ~\cite{DBLP:conf/iclr/Roh0WS21} introduced FairBatch, a bilevel optimization-based batch selection algorithm that adaptively chooses minibatch sizes to improve model fairness without requiring modifications to model training. 
Xu \textit{et al.} ~\cite{DBLP:conf/cvpr/XuHSLLHL021} proposed a false positive rate penalty loss to mitigate bias in face recognition by increasing the consistency of instance false positive rate. 
Lin \textit{et al.} ~\cite{DBLP:conf/eccv/LinKJ22} proposed FairGRAPE, a pruning method that minimizes the disproportionate impacts 
by calculating per-group weight importance and preserving relative between-group importance during network edge pruning. 
Park \textit{et al.} ~\cite{DBLP:conf/cvpr/ParkLLHKB22} proposed Fair Supervised Contrastive Loss for fair visual representation learning. 
This line of work aims at getting a fairer model by explicitly changing the training procedure.

\textit{Post-processing Methods}
Post-processing work ~\cite{DBLP:journals/csur/MehrabiMSLG21,DBLP:conf/cvpr/WangDXZCW022,DBLP:conf/aies/KimGZ19,DBLP:conf/icassp/LohiaRBSVP19} focused on adjusting model predictions based on specific fairness criteria after the model training, which often use a holdout set which is not involved during the training procedure. 
Lohia \textit{et al.} ~\cite{DBLP:conf/icassp/LohiaRBSVP19} introduced a technique that detects fairness bugs from  outputs and makes appropriate adjustments. 
Kim \textit{et al.} ~\cite{DBLP:conf/aies/KimGZ19} introduced a technique that creates a new classifier with equal accuracy across different protected attributes. 
Wang \textit{et al.}~\cite{DBLP:conf/cvpr/WangDXZCW022} presented an approach that learns to perturb input data,  
rendering the models incapable of recognizing fairness-related features. 

\subsection{Empirical Studies of Model Fairness}

Empirical studies are essential for evaluating the effectiveness and efficiency of fairness improvement methods. 
Additionally, it can inspire researchers to pursue further innovations in fairness improvement. Most previous studies~\cite{DBLP:journals/corr/abs-1905-05786,DBLP:conf/fat/FriedlerSVCHR19,DBLP:conf/sigsoft/BiswasR20,DBLP:conf/sigsoft/Zhang022} on fairness have focused on numerical or tabular inputs. 
However, fairness issues in image data are more complex. 
Wang \textit{et al}.~\cite{DBLP:conf/cvpr/WangQKGNHR20} conducted an analysis of computer vision models, specifically targeting adversarial training approaches, while Chen \textit{et al.}~\cite{chen2024fairness} studied fairness improvement methods that work with tabular inputs. In contrast, our research delves into 13 approaches working with image inputs from three major categories. Additionally, we use three distinct datasets and encompass all commonly-used performance metrics.

\section{Conclusion}
\label{sec:conclusion}

Deep Learning (DL) models have been widely-adopted in many applications, particularly in ethical-sensitive domains. As a result, ensuring the fairness of DL models has been a emerging research problem. Although many fairness improving methods have been proposed in recent years, there still lacks a systematic empirical study to comprehensively compare the performance of different approaches. To fill this gap, this paper conducted the first large-scale empirical study, where we comprehensively analyzed the performance of 13 state-of-the-art methods from multiple aspects, e.g., metrics, datasets, and tasks. Based on the results, we have summarized a set of findings and implications to promote future studies in this research area.

\section*{Data Availability}
For the sake of Open Science, we make the replication package with source code and the experimental results publicly available at:
\link{}.

\begin{acks}
This work was supported by the National Natural Science Foundation of China under Grant Nos. 62202324 and 62322208, and CCF Young Elite Scientists Sponsorship Program (by CAST).
\end{acks}

\normalem
\bibliographystyle{ACM-Reference-Format}
\bibliography{ref}

\end{document}